# Sensing and Control of Friction Mode for Contact Area Variable Surfaces (Friction-variable Surface Structure)

Seita NOJIRI, Akihiko YAMAGUCHI *Member IEEE*, Yosuke SUZUKI, *Member IEEE*, Tokuo TSUJI, *Member IEEE*, and Tetsuyou WATANABE, *Member IEEE*

*Abstract*— Robotic hands with soft surfaces can perform stable grasping, but the high friction of the soft surfaces makes it difficult to release objects, or to perform operations that require sliding. To solve this issue, we previously developed a contact area variable surface (CAVS), whose friction changed according to the load. However, only our fundamental results were previously presented, with detailed analyses not provided. In this study, we first investigated the CAVS friction anisotropy, and demonstrated that the longitudinal direction exhibited a larger ratio of friction change. Next, we proposed a 'sensible' CAVS, capable of providing a variable-friction mechanism, and tested its sensing and control systems in operations requiring switching between sliding and stable-grasping modes. Friction sensing was performed using an embedded camera, and we developed a gripper using the sensible CAVS, considering the CAVS friction anisotropy. In CAVS, the low-friction mode corresponds to a small grasping force, while the high-friction mode corresponds to a greater grasping force. Therefore, by controlling only the friction mode, the gripper mode can be set to either the sliding or stable-grasping mode. Based on this feature, a methodology for controlling the contact mode was constructed. We demonstrated a manipulation involving sliding and stable grasping, and thus verified the efficacy of the developed sensible CAVS.

## I. Introduction

Robotic hands with soft surfaces enable stable grasping, however it is difficult to slide the finger surfaces on an object surface for manipulating it. The main factors that contribute to grasping stability are elastic deformation, which helps the surface conform to object shapes, and the high level of friction achieved with many object types [1]. Many types of soft robotic hands have demonstrated their effectiveness in grasping different object types [2], however, such soft robotic hands have not been able to manipulate those objects very well [2]. In order to manipulate an object in hand, or to perform a task using the grasped object, it is necessary to control the grasped object in a position and posture suited to the purpose. However, unpredictable deformation in soft surfaces disturbs the motion of objects, and in such cases, sliding on soft surfaces will be effective in removing such disturbances or interference—and this sliding motion can also extend the range of available manipulations. An example of an operation involving sliding would be sliding fingers on the surface of a target object in order to reach contact positions appropriate for grasping and manipulation. However, soft surfaces exhibit high levels of friction with target objects, and it becomes difficult to perform the sliding operation necessary to manipulate the grasped object.

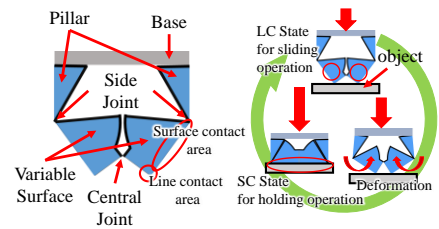

Fig. 1 Overview of a contact area variable surface (CAVS). SC: Surface contact, LC: Line contact

To address this issue. our group developed a contact area variable surface (CAVS), whose friction changed according to the load [3]. Friction on soft surfaces is largely associated with the (apparent) contact area, and thus we proposed the concept of changing friction by changing the contact area. The main features of CAVS are that it works on soft bodies passively, and exhibits low friction with a light touch and more friction with a heavy touch. The fundamental results from this work have been previously presented [3], with the detailed analyses, including the CAVS anisotropy, not yet provided. In addition, we did not use CAVS in robotic manipulations, and nor did we introduce any sensing or control methodology for the contact-surface state. An example of the type of operation that CAVS can be expected to perform effectively is one that requires switching between sliding and stable grasping (holding). For sliding across the surface on an object, the friction should be small, whereas, for stably grasping an object, the friction level should be high. This study covers the following topics:

**Frictional anisotropy of CAVS**: The structure of CAVS has anisotropy. Therefore, we investigate the frictional anisotropy of CAVS, and show that the longitudinal direction exhibits a larger ratio of friction change.

**Sensible CAVS**: We introduce a camera into CAVS, in order to sense the state of the CAVS internal structure. We

*Research supported by JSPS KAKENHI Grant Numbers 18K19809 and 18K18831 and the Cabinet Office (CAO), Cross-ministerial Strategic Innovation Promotion Program (SIP), "An intelligent knowledge processing infrastructure, integrating physical and virtual domains" (funding agency: New Energy and Industrial Technology Development Organization (NEDO)).

S. Nojiri is with the Graduated school of Natural science and Technology, Kanazawa University.

A. Yamaguchi is with the Graduate School of Information Sciences,Tohoku University, Sendai, Miyagi, 9808579, Japan.
Y. Suzuki, T. Tsuji, and T. Watanabe are with the Faculty of Frontier Engineering, Institute of Science and Engineering, Kanazawa University, Kakuma-machi, Kanazawa, 9201192 Japan (e-mail: te-watanabe@ieee.org).

create a model to identify the CAVS friction mode (low / high), from the internal structure state, which will be useful in applications involving object manipulation with CAVS-enabled robotic grippers.

**Controlling friction mode with a CAVS-enabled robotic gripper**: We develop a gripper using sensible CAVS, considering its investigated frictional anisotropy. We also present a methodology for controlling the friction mode of the gripper, and demonstrate a manipulation that involves sliding and gripping operations, to show the efficacy of the developed sensible CAVS.

*A. Related works*

There have been many studies on robot hands with flexible finger surfaces or entire flexible fingers [1], [2], [4]–[11]. For example, Ilievski et al. developed a pneumatically operable gripper with a flexible body [4], while Wang et al. succeeded in grasping several food items that could be broken easily, by using a pneumatically driven silicone gripper [5]–[7]. Hawkes et al. achieved shear grasping, with a gripper, using a gecko-inspired adhesive [8], and then Glick et al. combined this gecko adhesive with fluidic elastomer actuators to improve gripping ability [9]. Several grippers were developed based on a Fin-Ray structure, which passively bent and conformed to the curvature of the object being touched [10], [11]. In some studies, fluid fingers were created, from an elastic membrane filled with fluid, making it easier to adapt to the object shape and grasp fragile objects stably [12]–[14]. Mizushima et al. proposed a silicone texture for the surface of the fluid finger, and demonstrated that the surface provided high friction levels, even under wet conditions [15]. The jamming grippers provided soft-touch capability, by using an elastic membrane filled with granular material, and hard-holding capability, through a jamming transition created using air vacuuming [16]–[25]. However, these soft robotic hands were aimed at stable grasping and did not consider manipulation, particularly manipulation involving sliding.

Several studies have dealt with friction change on robot surfaces. One of the popular methodologies has been friction change achieved by changing the material at contact. Becker et al. developed a mechanism in which a sticky material was pushed out through holes on an outer cover by inflating an inner silicone balloon [26]. Spiers et al. developed a similar mechanism, with the difference being that the mechanism worked passively, according to the magnitude of the load [27]. Unlike these systems, CAVS can be constructed on soft bodies.

Changes in the shape or property of surface materials can also produce friction change. Suzuki et al. developed a silicone body with wrinkles on the surface, whose size and shape were controlled by compressing and stretching the body, to control friction [28]. Liu et al. also used changes in wrinkles to control friction, with the difference being that the control method was based on UV light exposure [29]. Kim et al. developed a material that could adjust its adsorptive power according to the temperature, using a shape memory polymer [30]. Abdi et al. considered a different method for controlling friction [31], in which they controlled skin shape by using capacitive, microelectromechanical system actuators. Shintake et al. develop a gripper whose electroadhesion force on the contact surface could be controlled by dielectric elastomer actuators [32], and Mizushima et al. developed a friction-reduction system for high-friction textures [15], by injecting lubricants. Unlike these systems, CAVS does not need additional systems or space for friction control. In addition, in CAVS, the low-friction mode corresponds to a small grasping force, whereas the high-friction mode corresponds to a greater grasping force. Therefore, by controlling the friction mode alone, the gripper mode can be changed between sliding or stable grasping modes. This benefit has been used in constructing a methodology for controlling the friction mode for manipulation.

In this work, a camera is installed in CAVS, for observing and controlling the friction mode. The small cameras, developed recently, have been used to obtain tactile information, by being installed in robotic fingers [33]–[42]. Generally, by using a camera to capture the displacements of the markers attached to the translucent elastic body, magnitude and distribution information of the force can be acquired [33]–[35]. When using a transparent elastic body as the surface [36], [37], both force–distribution information and slip detection and proximity information can be obtained. Cameras have not been used for sensing the friction-mode of surfaces whose friction is controllable, so far, however, and so in the work reported here, we extended this concept to observation of the friction mode. To demonstrate the usefulness of the sensible CAVS, we have developed a method for sensing and controlling the friction mode of the sensible CAVS, and tried to achieve manipulation that required switching between sliding and stable grasping (holding) modes.

The remainder of this paper is organized as follows: Firstly, an introduction to CAVS is provided, followed by investigation of CAVS frictional anisotropy, and a description of the proposed sensible CAVS that includes discussion of the sensing and control of the friction mode. Manipulation results for the proposed sensible CAVS are also provided.

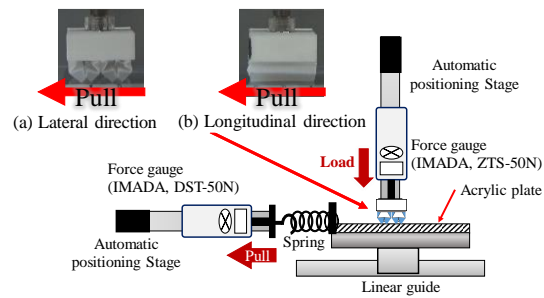

Fig. 2 Experimental setup for investigating frictional anisotropy of a contact area variable surface (CAVS)

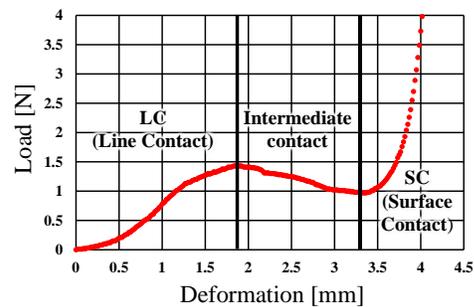

Fig. 3 The relationship between pressing force and deformation

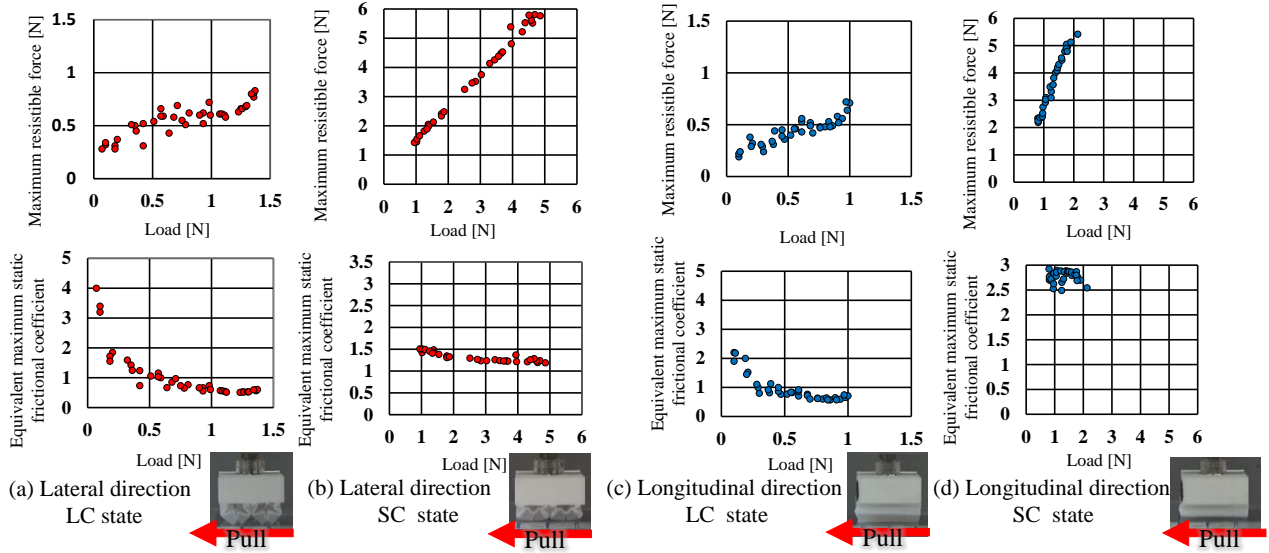

Fig. 4 Relationships between pressing force vs maximum resisitible force (tangential force value when sliding started) and equivalent coefficient of maximum static frictional force (ECMSF) derived by (1), at (a) line contact (LC) state in the lateral direction, (b) surface contact (SC) state in the lateral direction, (c) LC state in the longitudinal direction, and (d) SC state in the longitudinal direction

## II. OVERVIEW OF CAVS

Fig. 1 is a schematic of the CAVS structure and its friction-change mechanism. The CAVS consists of two variable surfaces and two pillars made of silicone (Dragon Skin 30); the pillar and variable surface are connected at the side joint, and the two variables surfaces are connected at the central joint. The inner surface of the CAVS (light blue areas in the left figure) is covered with a piece of cloth (100% cotton, HANDY CROWN CORPORATION) for enhancing structural rigidity. If the pressing load is small, CAVS contacts the object at the line contact (LC) area, which indicates low friction, and a sliding operation can be easily performed. If the load is large, the contact area changes to the surface contact (SC) area, as the LC area slides in and the central and side joints bend. A large contact area on soft surfaces indicates high friction levels, which is suitable for the stable holding of an object. Upon load removal, CAVS returns to its initial LC state. Refer to [3] for the other details, including the fabrication process.

## III. ANISOTROPY OF CAVS

As can be seen from the anisotropic structure of CAVS, its frictional property can vary, depending on the direction. Here we investigate this property, using the experimental setup shown in Fig. 2. The CAVS was attached to a force gauge (IMADA ZTS-50N), which was attached to an automatic stage (IMADA MX2-500N) used for moving the CAVS vertically and pressing it against the acrylic plate. The acrylic plate was attached to a linear guide, to allow the plate to move in the horizontal direction. The plate was connected to an automatic positioning stage (Oriental Motor ELSM2XF030K) with a force gauge (IMADA DST-50N), via a spring (MISUMI AWA5-20), to apply a tangential force at the contact area, and to measure the tangential force value. The spring was used to reduce the impact of sudden movement by the acrylic plate when the applied tangential force exceeded its limit, and sliding started. A video camera was used to record the CAVS contact state during the investigations.

First, we investigated the relationship between the pressing force and deformation amount, to identify the ranges of the LC and SC states. We pressed the CAVS against the plate, at the speed of 10 mm / min, to minimize dynamic effects. The obtained relationship between the ranges of the LC and SC states is shown in Fig. 3. The LC state finished at the local maximum (load: 1.43 N, deformation: 1.9 mm), and after that, a degree of deformation close to buckling relaxed the compression, and reduced the pressing load. The SC state started at the local minimum (load: 0.97 N, deformation: 3.3 mm), in which the entire SC area contacted the plate.

Based on these results, we investigated friction forces across LC and SC state ranges. The pressing force was initially set to a value within each range, and the set value was varied from the minimum to the maximum value, in steps of 0.1 N (LC state: 0.1–1.3 N, SC state: 1.1–2.0 N). After pressing the CAVS against the plate with the desired pressing force, we pulled the plate until sliding occurred (the plate started moving). We investigated pulling in lateral and longitudinal tangents (in [3], only the lateral direction was examined), and recorded the tangential force value when sliding started, referring to this as the maximum resistible force, $f_{MAX}$. This force is affected not only by frictional force, but also by elastic deformation, and so we evaluated the friction using the following equivalent coefficient for the maximum static frictional force (ECMSF):

$$\mu = \frac{f_{MAX}}{f_{Nslip}}. \qquad (1)$$

While maintaining the pressing force constant was difficult, and thus, we conducted the experiment while maintaining the pressing amount constant. In this case, the pressing force while applying a tangential force varied, so we therefore evaluated friction using $f_{Nslip}$—the pressing force value when sliding started. The obtained $f_{Nslip}$ could vary, and the time when the sliding started was detected by a change in the tangential force, and by examining the video camera footage. In this investigation, each setting was tested three times.

Fig. 4 shows our experimental results, and it can be seen that the proposed concept, in which the friction level was varied by changing the contact area, was effective in both directions. When the pressing force was small in the LC state, the ECMSF was large in both directions. The intercept of the relation of $f_{Nslip}$ vs $f_{MAX}$ was not the origin, because of the adhesion forces, and the non-zero intercept caused the large ECMSF seen for the small $f_{Nslip}$. Except for the range of small $f_{Nslip}$, the ECMSFs for both directions exhibited similar results—a constant small value in the LC state. In the SC state, the ECMSFs exhibited constant large values (over twice that of the LC state). When comparing the difference caused by using different pulling directions, the tendency to increase was the same, whereas the values were different. The ECMSF in the longitudinal direction was approximately twice that for the lateral direction, indicating that more stable grasping could be expected in the longitudinal direction.

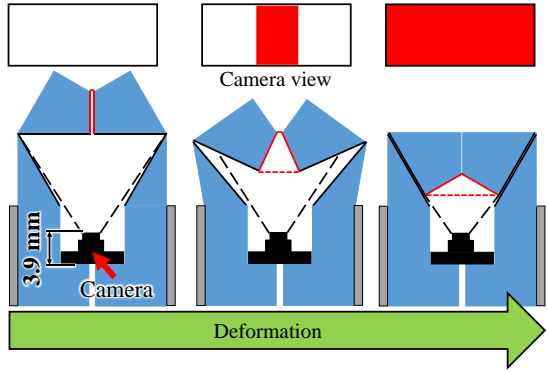

Fig. 5 Structure and principle of the sensing system

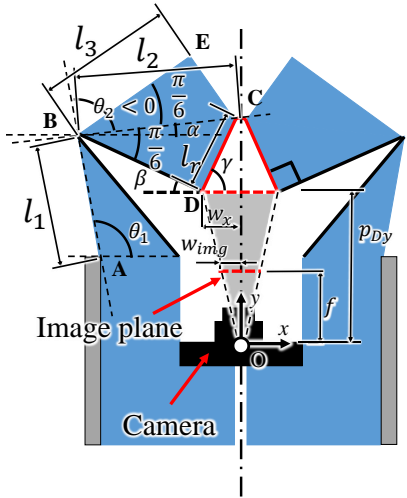

Fig. 6 Nomenclature for sensing the red area visible to the camera

## IV. SENSIBLE CAVS

To utilize the CAVS in practical operations, the friction mode must be observed. The deformation amount corresponds to the friction mode, and the surfaces around the central joint largely rotate with the deformation. In this section, we present a method for identifying the friction mode, by observing the change in the inner surface, using a camera embedded within the CAVS.

### A. Structure and principle of the sensing system

Fig. 5 shows the structure of the sensible CAVS. The surfaces around the central joint connecting the variable surfaces are colored red. The camera is located inside the CAVS, so that the red area can be captured by the camera, even when the CAVS has been deformed completely. The amount of red area visible to the camera changes with the amount of deformation, so by using the camera to detect change in the observable red area, we can estimate the deformation and the corresponding friction mode.

### B. Sensing friction mode from a camera image

Here, we describe how we can estimate the friction mode, or contact state, from the information on the red area that is visible to the camera. Fig. 6 shows the nomenclature used for this analysis. Here we will derive the relation between the deformation of CAVS, $d$, and the width of the captured red-colored part on the image plane of camera, $w_{\text{img}}$, in order to see how to control the $w_{\text{img}}$ and achieve the desired friction mode. It should be noted that the red area in the $z$ axis was painted red uniformly, so therefore, the red area observed by the camera should correspond to the width. We also neglect the deformation effect of the silicone parts in this analysis, by assuming that their deformation effect is small.

As shown in Fig. 6, the camera captures the red-colored part of the variable surface projected onto the camera's image plane. First, we derive the width of the captured red-colored part, $w_x$ that corresponds to $w_{\text{img}}$.

$$w_x = l_r \cos \gamma. \quad (2)$$

where

$$\begin{aligned} \alpha &= \theta_1 + \theta_2, \\ \beta &= \tfrac{\pi}{6} - \alpha = \tfrac{\pi}{6} - (\theta_1 + \theta_2), \\ \gamma &= \tfrac{\pi}{2} - \beta = \tfrac{\pi}{3} + \theta_1 + \theta_2. \end{aligned} \quad (3)$$

Let $\boldsymbol{p}_E = [p_{Ex} \ \ p_{Ey}]^{\mathrm{T}}$ be the position of point E, and $\boldsymbol{p}_A = [p_{Ax} \ \ p_{Ay}]^{\mathrm{T}}$ be the position of point A. Then, $\boldsymbol{p}_E$ can be represented by

$$\begin{bmatrix} p_{Ex} \\ p_{Ey} \end{bmatrix} = \begin{bmatrix} l_1 \cos \theta_1 + l_3 \cos(\theta_1 + \theta_2 + \pi/6) \\ l_1 \sin \theta_1 + l_3 \sin(\theta_1 + \theta_2 + \pi/6) \end{bmatrix} + \begin{bmatrix} p_{Ax} \\ p_{Ay} \end{bmatrix}. \quad (4)$$

Here, because $\boldsymbol{p}_A$ is constant, $\boldsymbol{p}_E$ is the function of $\theta_1$ and $\theta_2$. Because the deformation, $d$, corresponds to the displacement of $p_{Ey}$ from its initial position, it can be given by:

$$d = -(p_{Ey} - p_{Ey0}) \quad (5)$$

where $p_{Ey0}$ denotes $p_{Ey}$ at the initial position, where the CAVS in not in contact with anything. Here, the $x$ coordinate of the point C (Central joint) is constant, and is expressed as shown in (6):

$$p_{cx} = l_1 \cos\theta_1 + l_2 \cos(\theta_1 + \theta_2) + p_{Ax} = p_{cx0} \quad (6)$$

where $p_{Cx0}$ denotes $p_{Cx}$ at the initial position. If $d$ is given, the corresponding $\theta_1$ and $\theta_2$ can be derived from (4), (5), and (6), and then $w_x$ can be derived from (2), as shown in (7):

$$w_x = w_x(\theta_1, \theta_2) = w_x(d). \quad (7)$$

On the other hand, the width of the red-colored part on the image plane, $w_{img}$, is given by:

$$w_{img} = \frac{f}{p_{Dy}} w_x, \quad (8)$$

where $f$ denotes the focal length, and $p_{Dy}$ is the $y$ coordinate of the point D, and is given by:

$$p_{Dy} = l_1 \cos\theta_1 + l_3 \cos(\theta_1 + \theta_2 - \pi/6) + p_{Ay} \quad (9)$$

Therefore, $p_{Dy}$ is the function of $\theta_1$ and $\theta_2$, namely $d$. If summarizing, $w_{img}$ is the function of $d$ from (7), (8), and (9), as shown in (10).

$$w_{img} = \frac{f}{p_{Dy}(d)} w_x(d) \quad (10)$$

We evaluate $w_{img}$ by normalizing, using the value when the CAVS reached the SC state, $w_{SCimg}$, where $d = d_{SC}$, to remove $f$ —which is obtained as a pixel value by camera calibration, and cannot be easily treated due to the difference in units.

$$r_{img} = \frac{w_{img}}{w_{SCimg}} = \frac{p_{Dy}(d_{SC})}{p_{Dy}(d)} \frac{w_x(d)}{w_x(d_{SC})}. \quad (11)$$

In the calculation of the ratio $r_{img}$, parameter values are required (see Table I). Camera calibration is required to obtain $p_{Ay}$, while $d_{SC}$ is obtained from the experimental results (Fig. 3), and the other parameter values are the design parameters that can be obtained from the CAVS design.

The relationship between $r_{img}$ and $d$ was calculated, based on (11). The results are shown in Fig. 7, indicating the experimentally obtained values when the CAVS was pressed against the acrylic plate with the setup shown in Fig. 2. Note that the ranges for both contact modes were slightly different from those in Fig. 3, due to individual differences in the CAVS. From Fig. 7, it can be seen that the calculation results matched the experimental data very well, and that the methodology for deriving $r_{img}$ using (11) (including the assumption that the deformation effect of the silicone parts was negligible) was validated. If the $r_{img}$ is below 44%, the contact state corresponds to the LC state. Therefore, if the size of the red area at the SC state has been captured by the camera, we can estimate the state by checking whether the detected red area is within 44% of the obtained red area for the SC state. The friction mode of the CAVS can be easily controlled by controlling the deformation amount so that the $r_{img}$ can be within 44% for the LC state and 100 % for the SC state.

TABLE I. PARAMETER VALUES FOR DERIVING THE WIDTH OF RED AREA VISIBLE BY CAMERA

| Parameter | Value |
|---|---|
| $d_{SC}$ | 3.5 [mm] |
| $p_{Ax}, p_{Ay}$ | $-5$, 2.72 [mm] |
| $p_{Cx0}$ | 0 [mm] |
| $p_{Ex0}, p_{Ey0}$ | 2.5, 8.66 $\left(= 5\sin\frac{\pi}{3} \times 2\right)$ [mm] |
| $l_1, l_2, l_3$ | 4.33 $\left(= 5\sin\frac{\pi}{3}\right)$, 5.77 $\left(= 5/\cos\frac{\pi}{6}\right)$, 5.0 [mm] |

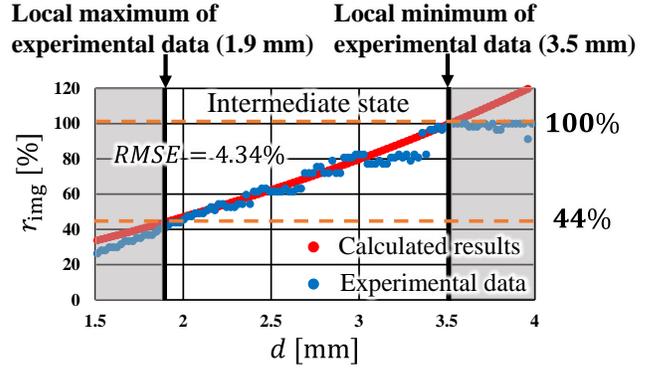

Fig. 7 Relationship between the calculated results and the red area visible by the camera

## V. UTILIZING SENSIBLE CAVS FOR MANIPULATION

In order to verify the effectiveness of the developed sensible CAVS, we conducted handling experiments with them, using a tube manipulation task requiring both sliding and holding operations. The manipulation test involved training a tube around a cylindrical object, and then attaching the hook at the tip of the tube to a pin. The flexibility of the tube makes the operation difficult. Once the tube is released, it can deform easily and re-grasping is difficult. We addressed this issue by sliding while maintaining contact with the tube.

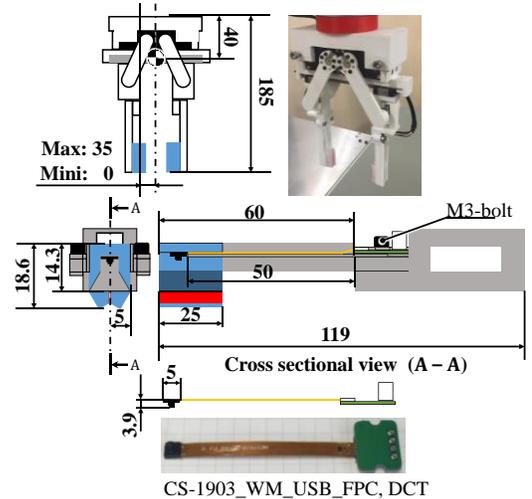

Fig. 8 CAVS-enabled robotic gripper

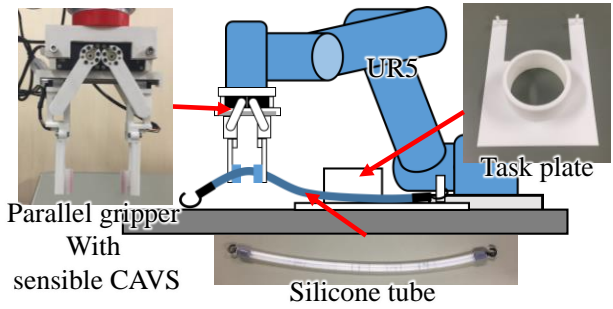

Fig. 9 Experimental setup for tube manipulation

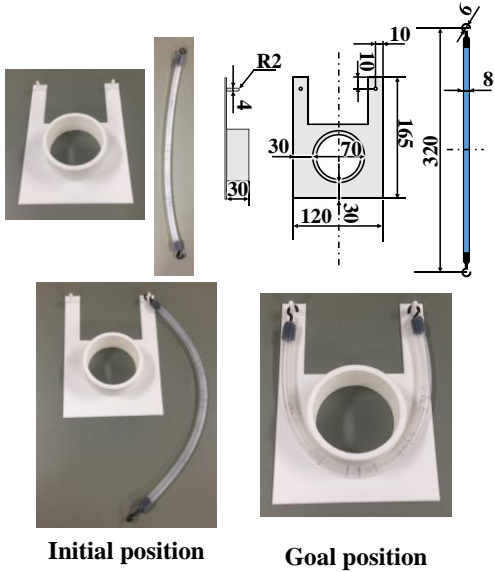

**Initial position**   **Goal position**

Fig. 10 Target object (silicone tube) and task plate

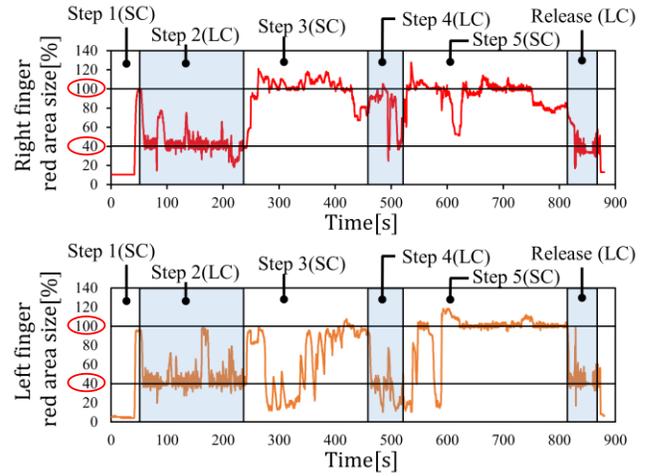

Fig. 11 Time-series data of the red area and corresponding (desired) contact state during the test manipulation

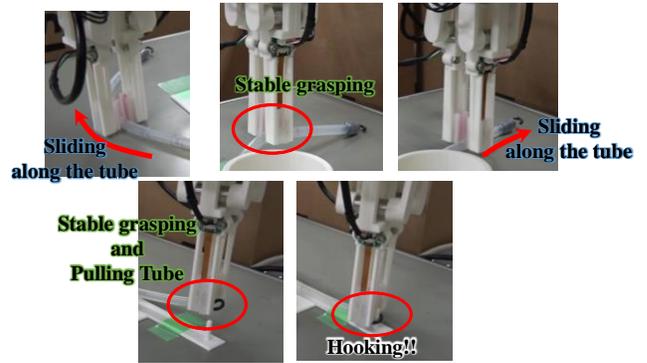

Fig.12 Overview of the manipulation

## A. CAVS-enabled robotic gripper

The parallel gripper shown in Fig. 8 was prepared for the manipulation, sensible CAVS was attached to each finger of the gripper, and the length in the longitudinal direction of the CAVS was set to 25 mm. As shown in Fig. 4, the longitudinal direction exhibited a larger ratio of friction change. Therefore, the finger was constructed so that the longitudinal direction of the finger could correspond to the longitudinal direction of CAVS—as the longitudinal direction of the finger corresponded to the gravitational direction for most of the operation.

## B. Experimental setup

The experimental setup is shown in Fig. 9. We used a robotic arm (UR5) equipped with the gripper, with sensible CAVS at the tip. As shown in Fig. 10, we prepared the task plate containing a cylinder (diameter: 70 mm, height: 30 mm), and two pins (diameter: 4 mm, height: 17 mm) for hooking. The target object was a tube with hooks attached to either end (Fig. 10). Its diameter was 8 mm and its length was 320 mm. Initially, one hook was attached to the pin on one side. The task was for the robot system to pick up the tube, train it around the cylinder, and attach the hook on the other end of the tube to the pin on the other side of the cylinder.

## C. Procedure

The main procedure for the test was as follows:

Step 1. Grasp the tube

Step 2. Slide the fingertips along the tube to the position required for the subsequent operation

Step 3. Wrap the tube around the cylindrical object while holding the tube

Step 4. Slide the fingertips along the tube to the position required for the subsequent operation

Step 5. Attach the hook to the pin while holding the tube.

Initially, the robot grasps the tube in the SC state, and then we controlled the contact state based on the desired state that we required. The LC state was selected as the desired state for sliding, while, the SC state was selected as the desired state for holding. The desired contact states were switched in each step. In Step 2, the desired positions for the wrapping operation were set at 20 mm from the hook. For the wrapping operation in Step 3, we controlled the position and posture of the gripper, so that the silicone tube could maintain contact with the cylinder. In Step 4, sliding was performed while maintaining contact between the tube and the cylinder, to preserve the wrapping achieved to date. The desired positions were around the hook for the subsequent hooking operation. The hooking operation in Step 5 required stretching the tube, and was suitable for evaluating whether the SC state could effectively retain its stable holding state, even when a disturbing force was exerted. We controlled the position and orientation of the gripper while stretching the tube.

Control of the gripper width for controlling the contact state was performed automatically, by setting the desired red area at 40% for the LC state, based on the results shown in Fig. 7 (taking account of the noise effect), and 100% for the SC state. Note that the 100% corresponds to the observed red area when controlling the contact state to the SC state at calibration. Gripper width was controlled by controlling the finger position so that the red area could be kept at its desired value, as shown in (12):

$$\begin{cases} \Delta d_\text{f} = \Delta d_{\text{fopen},i} \ (r_\text{img} - r_\text{imgd} > \varepsilon) \\ \Delta d_\text{f} = 0 \ (|r_\text{img} - r_\text{imgd}| \leq \varepsilon) \\ \Delta d_\text{f} = \Delta d_{\text{fclose},i} \ (r_\text{img} - r_\text{imgd} < -\varepsilon) \end{cases} \quad (12)$$

$$i \in \{left, right\}$$

where $\Delta d_\text{f}$ denotes the desired displacement for the finger position, $\Delta d_{\text{fopen},i}$ and $\Delta d_{\text{fclose},i}$ are the $\Delta d_\text{f}$ for opening and closing the finger respectively where the subscript $i$ denotes the left or right finger, $r_\text{imgd}$ denotes the desired $r_\text{img}$, and $\varepsilon$ is a positive constant (0.5). The $r_\text{imgd}$ was 40% for the LC state while 100% for the SC state. We set that $\Delta d_{\text{fopen},i} = 0.25$ mm and $\Delta d_{\text{fclose},i} = -0.5$ mm where the opening direction is positive. These values were selected by try and error. The reason of $|\Delta d_{\text{fclose},i}| > \Delta d_{\text{fopen},i}$ and of constant $\Delta d_{\text{fclose},i}$ and $\Delta d_{\text{fopen},i}$ is for reducing the risk of detaching from the object (tube) in the opening motion.

The other operations were conducted by manually controlling the robotic arm.

*D. Results*

Fig. 11 shows the time-series data of the red area, and the corresponding (desired) contact state during the manipulation, while Fig. 12 presents an overview of the operations. From Fig. 12, it can be seen that the given manipulation was completed without dropping or releasing the tube, while from Fig. 11, it can be seen that the red area was maintained at around its desired value—and that the corresponding desired contact state was achieved.

Note that the red area value was slightly adrift from the desired state several times. Changes in the position and orientation of the tube sometimes disturbed the contact states, causing disturbing forces, and automatic adjustment of the gripper width was thus conducted. It should also be noted that there were some cases in which the red area value exceeded 100%, due to excessive load (compared to the load corresponding to the SC state at the setting).

In Steps 1 and 2, contact area control was stable. In Step 3, the wrapping operation required a large change in the position and orientation of the gripper, and so control of the red area was slightly less stable than it had been in Steps 1 and 2. The sliding operation in Step 4 was conducted while the tube was under pressure or force, rendering control of the red area slightly less stable than it had been in Steps 1 and 2 —although the sliding state was still maintained during the operation. In Step 5, the red area remained stable, as did the grasping. At 750s, the red area of the right finger was lowered slightly, due to the contact between the pin and the hook, which caused the red area to fluctuate—although the red area of the left finger was not disturbed at this time, indicating that grasping has been maintained.

After that, the hook was attached to the pin, while holding the tube in the SC state. After Step 5, we kept the contact state in the LC state, to release the fingers from the tube. The results indicated that the developed sensible CAVS system worked effectively for manipulation involving both sliding and holding, through appropriate switching of the contact state.

VI. CONCLUSION

In this paper we have presented the results of work involving a novel sensible CAVS, providing a variable friction mechanism and a sensing and control system, for operations requiring switching between sliding and stable grasping. The variable friction mechanism was provided by the change in the contact area stemming from deformation. In addition, a camera was embedded into the CAVS, to estimate the contact state, and, using the resultant visual information, the method for estimating the contact or friction mode was provided. Using the estimated contact or friction-mode information, we presented a control method in which friction was controlled in the contact area, to create contact states suited to either sliding or holding operations.

The validity of the sensible CAVS and its control system was demonstrated by a tube manipulation test, which required switching between sliding and stable grasping. Although the concept of CAVS was provided previously, the detailed analyses presented here had not been provided. Therefore, in this work, we also investigated the directional property of friction in CAVS, showing that the longitudinal direction exhibited a larger ratio of friction change. This result was then applied to construction of the gripper.

The method for estimating the contact or friction mode could be extended to the method for estimating the contact force. If a transparent material is used for constructing CAVS, the embedded camera can capture much more information, including contact sliding detection or proximal information, as well as various other types of tactile information. CAVS size has not yet been optimized, however, and CAVS will need to be miniaturized to deal with small objects. These issues are beyond the scope of this study, and will be covered in our future work.